%% file: 00-main.tex
\pdfoutput=1

\documentclass[11pt]{article}
\usepackage{hyperref}
\usepackage{xspace}\usepackage[dvipsnames]{xcolor}
\usepackage{booktabs}
\usepackage{float}
\usepackage{graphicx}
\usepackage{caption}
\usepackage[]{acl}

\usepackage{tabularx}

\usepackage{times}
\usepackage{latexsym}

\usepackage{multirow}
\usepackage[T1]{fontenc}

\usepackage[utf8]{inputenc}

\usepackage{microtype}

\usepackage{inconsolata}
\usepackage{amsmath}
\usepackage{color, colortbl}
\definecolor{gainsboro}{rgb}{0.86, 0.86, 0.86}

\usepackage[normalem]{ulem}

\newcommand{\methodname}{\textsc{MaRCo}\xspace}
\newcommand{\dexperts}{\textsc{DExperts}\xspace}

\title{
Detoxifying Text with \methodname:\\Controllable Revision with Experts and Anti-Experts
}

\newcommand{\aspace}{\hspace{1.2em}}
\newcommand{\uw}{$^{\heartsuit}$}
\newcommand{\cmu}{$^{\diamondsuit}$}
\newcommand{\aiTwo}{$^{\clubsuit}$}
\author{
    Skyler Hallinan\uw\aspace 
    Alisa Liu\uw\aspace
    Yejin Choi\uw\aiTwo \aspace
    Maarten Sap\cmu\aiTwo\aspace \\
    \small{\uw Paul G.\ Allen School of Computer Science \& Engineering, University of Washington} \vspace{-.1em}\\
    \small{\aiTwo Allen Institute for AI} \aspace
    \small{\cmu Language Technologies Institute, Carnegie Mellon University}\vspace{.1em}\\
    \texttt{\href{mailto:hallisky@uw.edu}{hallisky@uw.edu}}, \texttt{\href{mailto:maartensap@cmu.edu}{maartensap@cmu.edu}}
}

\begin{document}
\maketitle
\begin{abstract}
\input{sections/00-abstract}
\end{abstract}

\input{sections/01-introduction}

\input{sections/02-methods}

\input{sections/03-out-of-context}
\input{sections/06-conclusion}
\input{sections/07-limitations-ethics}




\bibliography{00-refs}

\appendix
\input{sections/10-appendix}

\end{document}

%% file: sections/00-abstract.tex





Text detoxification has the potential to mitigate the harms of toxicity by rephrasing text to remove offensive meaning, but subtle toxicity remains challenging to tackle. 
We introduce \methodname, a detoxification algorithm that combines controllable generation and text rewriting methods using a Product of Experts 
with autoencoder language models (LMs).
\methodname uses likelihoods under a non-toxic LM (expert) and a toxic LM (anti-expert) to find candidate words to mask and replace.
We evaluate our method on several subtle toxicity and microaggressions datasets, and show that it not only outperforms baselines on automatic metrics, but \methodname's rewrites are preferred 2.1$\times$ more in human evaluation. 
Its applicability to instances of subtle toxicity is especially promising, demonstrating a path forward for addressing increasingly elusive online hate.

\input{tables/fig1}




%% file: tables/fig1.tex
\begin{figure}[t!]
\centering
\includegraphics[width=0.9\columnwidth]{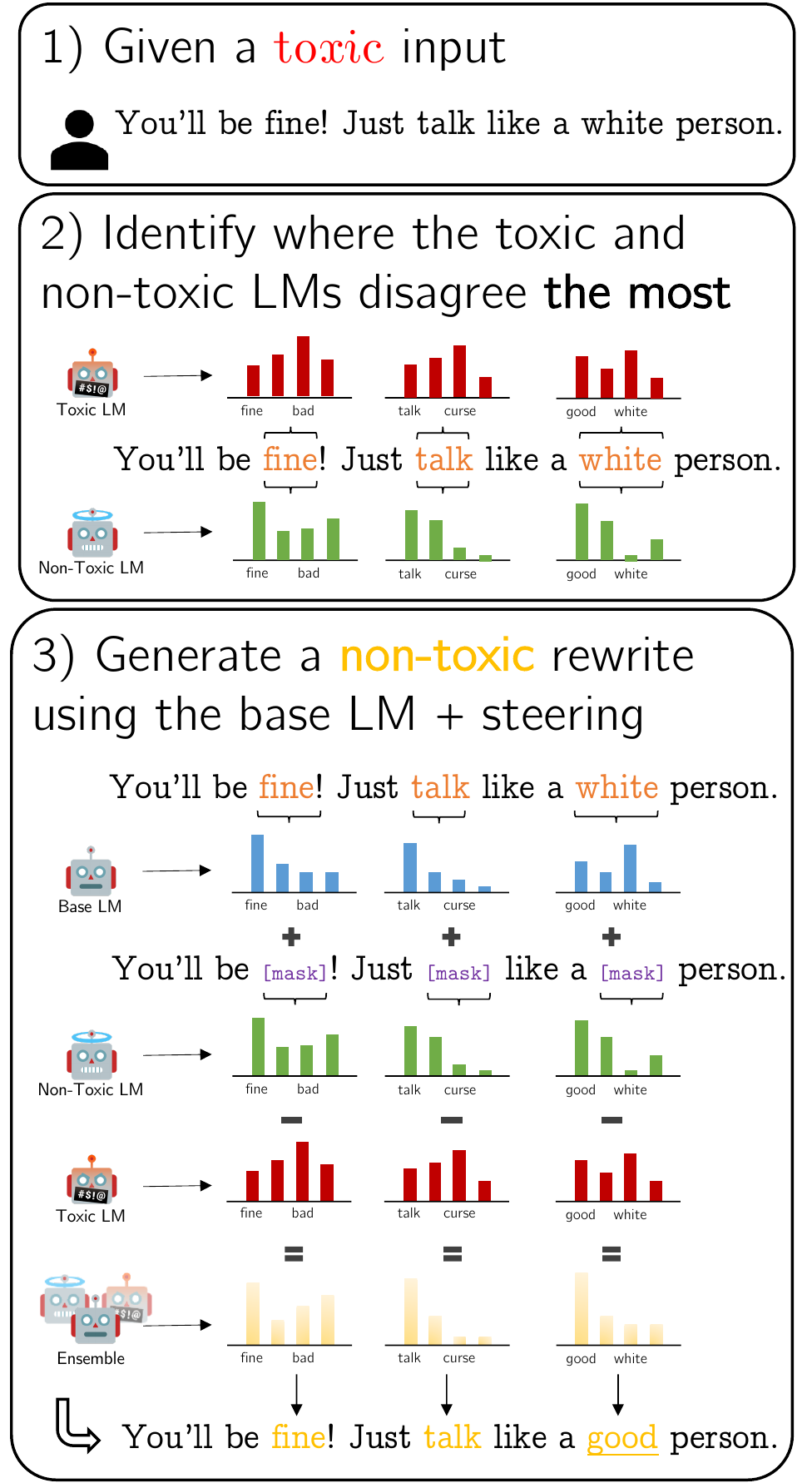}
\caption{A demonstration of the \methodname algorithm, which utilizes a base language model (LM) and a finetuned toxic and non-toxic LM to rewrite toxic text. We start with toxic text, identify potentially toxic tokens via disagreement of the toxic and non-toxic LMs, and finally generate a non-toxic rewrite using the base model steered by the toxic and non-toxic LM.}
\label{fig:fig1}
\end{figure}

%% file: sections/01-introduction.tex


\section{Introduction}
Toxic, offensive, hateful, or biased language is increasingly prevalent and can cause online and offline harms, especially to minority groups \cite{9519435,Ohchr2021-ba}.
This is challenging for NLP systems to detect and account for when biases are subtle or without explicit toxic keywords
\cite{https://doi.org/10.48550/arxiv.2203.09509, han-tsvetkov-2020-fortifying, vidgen-etal-2021-learning}.
For example, the statement "\textit{You'll be fine! Just talk like a white person}" conveys the biased implication that non-white dialects are not conducive to success (\autoref{fig:fig1}), which is a harmful racial stereotype \cite{https://doi.org/10.1002/j.1556-6676.2014.00130.x}.

\textit{Text detoxification}, i.e., rewriting text to be less toxic while preserving non-toxic meaning, provides a promising solution by suggesting alternative ways of expressing similar ideas with less biased implications \cite{nogueira-dos-santos-etal-2018-fighting}.
For example, the rewrite ``\textit{You'll be fine! Just talk like a \underline{good} person}" eliminates the racial bias from the original statement while preserving the \emph{non-toxic} meaning.
Such methods have the potential to improve the quality of online conversations \cite[e.g., through machine-in-the-loop interfaces;][]{Hohenstein2021-mn,10.1145/3172944.3172983}.



We present \methodname, \textbf{Ma}sk and \textbf{R}eplace with \textbf{Co}ntext: a new, unsupervised algorithm for text detoxification that combines mask-and-replace text denoising with controllable text generation using a Product of Experts (PoE) \cite[PoE, \dexperts;][]{10.1162/089976602760128018, liu-etal-2021-dexperts}.

\methodname jointly uses an expert and an anti-expert, a pair of language models (LM) fine-tuned on a \textbf{non-toxic} and \textbf{toxic} corpus respectively, to identify which tokens \emph{most likely} contribute to the overall toxicity, and then suggest replacements that lower toxicity.
Using LMs to capture toxicity allows \methodname to rewrite much subtler toxic text compared to previous work that uses toxicity classifiers or toxic word lists \cite{dale-etal-2021-text}.


We apply \methodname to three datasets focused on subtly toxic statements, such as microaggressions. 
Our method outperforms state-of-the-art detoxification baselines from \citet{dale-etal-2021-text} across all three datasets, as measured through both automatic and human evaluation. 
Our work shows the effectiveness of combining controllable generation with text rewriting methods
for text detoxification.\footnote{We release our code and data at \url{https://github.com/shallinan1/MarcoDetoxification}.}

%% file: sections/02-methods.tex
\section{Background: Text Detoxification}

\input{tables/all_auto}
Text detoxification is a form of stylistic rewriting \cite{10.5555/3305381.3305545, 10.5555/3295222.3295427, jhamtani-etal-2017-shakespearizing} with the goal of producing a non-toxic rewrite given a toxic input sentence. 
This task is challenging, as it requires both detoxification \textit{and} preservation of non-toxic meaning, in contrast to controllable text generation, which aims to simply generate \textit{any} non-toxic continuation for a prompt \cite{prabhumoye-etal-2020-exploring,gehman-etal-2020-realtoxicityprompts}.


Due to a lack of supervision with parallel data, an often effective approach to stylistic rewriting relies on unsupervised masking-and-reconstructing approaches \cite{li-etal-2018-delete,ijcai2019-732, malmi-etal-2020-unsupervised,ma-etal-2020-powertransformer}. In this paradigm, source style-specific tokens/spans in the input text are detected and masked, then filled in with tokens/spans from the target-style using a masked language model. 
Other work has framed detoxification as a translation or paraphrasing task, using a classifier to steer away from toxic content \cite{ nogueira-dos-santos-etal-2018-fighting, dale-etal-2021-text}.

\section{Text Detoxification with \methodname}


\methodname is an unsupervised approach to text detoxification, consisting of two discrete steps: \textbf{masking} 
and then \textbf{replacing} tokens, assisted by the \emph{context} of the entire sequence.
Though inspired by \dexperts \cite{liu-etal-2021-dexperts}, our novelty is two-fold: first, we tackle a more challenging task, unsupervised revision, instead of style-controlled generation, and second, we propose a \emph{detect} and \emph{rewrite} pipeline, in contrast to simple word-distribution steering during autoregressive generation.


\paragraph{Expert and Anti-Expert LMs}
Our method for unsupervised controlled revision is based on \textit{denoising autoencoder} LMs (AE-LMs), which are trained to mask and reconstruct sequences of text.
Our setup consists of
a \emph{base} pretrained AE-LM $\boldsymbol{G}$, an \emph{expert} AE-LM $\boldsymbol{G^+}$ finetuned on data with desirable attributes, and an \emph{anti-expert} AE-LM $\boldsymbol{G^-}$ finetuned on data with undesirable attributes. 

We use BART-base \citep{lewis-etal-2020-bart} as our base autoencoder. 
We finetune the expert and anti-expert using 1M non-toxic and 100K overtly toxic comments from the Jigsaw corpus \cite{Do2019JigsawUB}, as done in \citet{liu-etal-2021-dexperts} and \citet{dale-etal-2021-text}. 
BART can infill multiple or no tokens even if only one token is masked, allowing for more flexible mask infilling. 
See \autoref{modelingdetails} for training details. 








\subsection{Contextual Masking} 

We first identify locations that \emph{could} convey toxic meaning; intuitively, these could be words or phrases with strongly differing likelihoods under the expert and anti-expert.

Formally, given a sequence $\boldsymbol{w}$, for every token $w_i \in \boldsymbol{w}$, we temporarily mask it and generate probability distributions over the vocabulary $\mathcal{V}$ for that location from $G^+$ and $G^-$, 
which we denote $P^+$ and $P^-$ respectively. 
Then, we compute the distance $d_i$ between $P^+$ and $P^-$ using the Jensen-Shannon divergence, a symmetric form of the Kullback–Leibler (KL) divergence:\footnote{Given probability distributions $A$ and $B$, the KL divergence is defined as $
D_{\mathrm{KL}}(A \| B)=\sum\limits_{x \in \mathcal{V}} A(x) \log \left(\frac{A(x)}{B(x)}\right)$}
\begin{gather}
\scalebox{0.92}{$
\begin{align*}
d_i = \frac{1}{2} \left (D_{\mathrm{KL}}(P^+ \| P^-) \right) + \frac{1}{2} \left (D_{\mathrm{KL}}(P^- \| P^+) \right)
\end{align*}
$}
\end{gather}
After normalizing all distances by the mean, we mask all $w_i$ whose distance $d_i$ is above a threshold $\tau$ and denote the resulting sequence $\boldsymbol{w^{m}}$; 
these masked tokens are locations where toxicity \emph{may} be present due to expert and anti-expert disagreement.




\subsection{Contextual Replacing}
After masking potentially toxic locations, \methodname then replaces them with more benign tokens -- if they are indeed toxic -- to autoregressively produce a rewrite $\boldsymbol{g}$ given the original and masked sentences $\boldsymbol{w}$ and $\boldsymbol{w}^m$. We transform the \dexperts \cite{liu-etal-2021-dexperts} framework, which leverages a PoE to steer a model away from toxic generations by ensembling token probabilities, to enable rewriting by using AE-LMs.

We obtain the next-token unnormalized log-probabilities (i.e., logits) $z_i$, $z_i^+$, and $z_i^-$ from the base and expert AE-LMs $G$, $G^+$, and $G^-$, respectively, 
conditioned on the previously generated tokens $\boldsymbol{g_{<i}}$, the original sequence $\boldsymbol{w}$, and the masked variant $\boldsymbol{w}^m$. We then ensemble those logits into a modified next-token probability distribution:
\begin{gather}
\scalebox{0.92}{$
\begin{align*}
    P(X_i| \ \boldsymbol{g_{<i}},\boldsymbol{w}, \boldsymbol{w^m}) = \text{softmax}(z_i + \alpha_1 z_i^+ - \alpha_2 z_i^-) 
\end{align*} 
$}
\end{gather}
where $X_i$ is a random variable over the vocabulary $\mathcal{V}$ representing the next token at index $i$ given the previous generation $g_{<i}$, and our two hyperparameters $\alpha_1$ and $\alpha_2$ independently control the impact of the expert and anti-expert for more flexibility.\footnote{\autoref{poe} gives further intuition into understanding this equation as a PoE.} 
In our method, the expert and anti-expert use the masked sequence $\boldsymbol{w_m}$ as their input, while the base model uses the unmasked $\boldsymbol{w}$.
Intuitively, the base model tries to replicate the input sequence but is steered by an expert and anti-expert with contrasting probability distributions at the masked locations. 
This enables rewrites with minimal but meaningful edits on toxic tokens and preservation of non-toxic content.
Note that for a masked location, when the base model agrees more with the anti-expert than with the expert, the original token is most likely toxic and will be replaced in the rewrite. 
On the other hand, if the differences between the expert and anti-expert are not enough to sway the base model, the original token is most likely non-toxic and will be re-added in the rewrite.

%% file: tables/all_auto.tex
\begin{table*}[h]
\footnotesize
\centering
\begin{tabular}{@{}llrrrrrr@{}}
\toprule
          & & \multicolumn{3}{c}{Validation}    & \multicolumn{3}{c}{Test}        \\
          \cmidrule(lr){3-5} \cmidrule(lr){6-8} 
          & Method & Toxicity ($\downarrow$) & BERTScore ($\uparrow$) & Fluency ($\downarrow$) &Toxicity ($\downarrow$) & BERTScore ($\uparrow$) & Fluency ($\downarrow$)
          \\ \midrule
\multirow{4}{*}{MAgr} & \emph{Original} & 0.286 	 & 	-- &         51.49   & 0.272  &-- & 70.20     \\
& CondBERT &   \underline{0.161}	& \textbf{0.966} &	\underline{104.10}      & \underline{0.148} & \textbf{0.964} &  \underline{88.69}                 \\
& ParaGeDi  &     0.162	& 0.931  &	104.46     & 0.172 & 0.929  & 120.78      \\
& \methodname     &      \textbf{0.145} & \underline{0.958} &	\textbf{43.54}    & \textbf{0.141} & \underline{0.954} &  \textbf{39.10} \\
\midrule
\multirow{4}{*}{SBF}& \emph{Original} &   0.351	& --	 &58.46   & 0.344 &-- &  88.79   \\
&CondBERT &   \underline{0.202}	&	\textbf{0.961} & \underline{69.51}            & \underline{0.190} & \textbf{0.961}& 131.12     \\
&ParaGeDi  &     0.186 &		0.921 &	179.88 & 0.192 & 0.923  & \underline{99.96}       \\
& \methodname     &     \textbf{0.176}	&	\underline{0.947}	 &	\textbf{54.86}    & \textbf{0.186} & \underline{0.946}  & \textbf{48.75}  \\ 
\midrule
\multirow{4}{*}{\begin{tabular}{@{}c@{}}Dyna \\ Hate\end{tabular}} & \emph{Original} &   0.563		 & 	-- &	205.73  & 0.578  & -- & 220.42  \\
& CondBERT &   \underline{0.288}	&	\textbf{0.954}	& \underline{190.51}      & \underline{0.293} & \textbf{0.950} & \underline{200.20}   \\
& ParaGeDi  &     0.332 &	 0.918 &217.78  & 0.323 & 0.912 & 240.17         \\
& \methodname    &     \textbf{0.274}	&	\underline{0.939} &	\textbf{110.50}     & \textbf{0.277} & \underline{0.936}  & \textbf{128.84}    \\
\bottomrule
\end{tabular}
\caption{Automatic evaluations on detoxified generations on MAgr, SBF, and DynaHate for \methodname, ParaGeDi and CondBERT across all datasets and splits, \methodname achieves the lowest toxicity, best fluency, and second-best BERTScore, while CondBERT achieves the highest BERTScore. \textbf{Bold} indicates the best metric, and \underline{underline} indicates the second-best metric in each column for each dataset.} 


\label{tab:auto}
\end{table*}

%% file: sections/03-out-of-context.tex
\section{Detoxification Experiments \& Results}
In our experiments, we focus on rewriting sentences from three toxicity datasets, and use both automatic and human evaluations to measure \methodname's performance at detoxifying text.

\subsection{Datasets}

We seek to rewrite English sentences that are already known to be or annotated as toxic, especially sentences that contain more subtle or implicit biases (e.g., without swearwords). 
In contrast to the Jigsaw corpus used to finetune our experts, we use three out-of-domain datasets with subtle toxicity:

\paragraph{Microagressions.com} (MAgr) is a publicly available Tumblr blog where users can anonymously post about socially-biased interactions and utterances in the wild. Each post includes an offending {quote} and/or a {description} of the incident. 
We scrape all \emph{quotes}, resulting in a set of real-world microagression utterances. The validation and test set sizes are 238 and 298 respectively.
 
 
\paragraph{Social Bias Frames} \citep[SBF;][]{sap-etal-2020-social} is a corpus of socially biased and offensive content from various online sources. 
We use a subset of SBF from the microaggressions subreddit,\footnote{A subreddit is a topic-focused community on \href{https://www.reddit.com/}{Reddit}}
which contains subtly biased content \cite{breitfeller-etal-2019-finding}.
We use all posts where the majority of annotators marked the text as offensive. The validation and test set sizes are 92 and 114 respectively.

\paragraph{DynaHate} \cite{vidgen-etal-2021-learning} is an adversarially collected set of hate speech, where human annotators create examples that an iteratively improved hate-speech classifier cannot detect. We utilize all four rounds of hate-speech data and use all examples marked as hateful. The validation and test set sizes are 1,858 and 2,011 respectively.
\input{tables/human_eval}
\subsection{Baselines}
We compare \methodname to the two baseline approaches from \citet{dale-etal-2021-text}, which have shown state-of-the-art detoxification performance. See \autoref{experimentaldetails} for generation details.

\paragraph{ParaGeDi}
utilizes a class-conditioned language model (using control codes for toxic and non-toxic styles) on top of a paraphrasing language model to steer generated text towards a specific attribute. 

\paragraph{CondBERT}
follows a pointwise editing setup, first identifying tokens to mask in the input, then using a mask-filling model to replace them. In contrast to \methodname, CondBERT uses a lexicon-based approach to masking words by using weights from a whole-word, toxic language logistic classifier.  



\subsection{Evaluation Setup}
We perform automatic and human evaluations, following previous work.

\paragraph{Automatic Metrics}
We assess the quality of the models' rewrites with automatic metrics used in previous work \cite{liu-etal-2021-dexperts,ma-etal-2020-powertransformer}.
We report the average \textbf{toxicity} score of rewrites using the PerspectiveAPI.\footnote{\url{www.perspectiveapi.org}, accessed 06-2022.} 
Additionally, we measure \textbf{fluency} of rewrites by computing their perplexity with an external LM \cite[GPT-2 XL;][]{radford2019language}, and \textbf{meaning similarity} between the input and the rewrite using BERTScore \citep{DBLP:journals/corr/abs-1904-09675}.
See Appendix \ref{evalmetrics} for further details.
\input{tables/onegen}

\paragraph{Human Evaluation}

We conduct a head-to-head human evaluation \cite{kiritchenko-mohammad-2017-best} of the toxicity of the rewrites using Amazon Mechanical Turk. 
For each dataset's validation and test sets, we sample 75 prompts each, then compare each pair of \methodname, ParaGeDi and CondBERT's generations against each other and ask which one is less toxic (along with an option to flag either of the rewrites as ungrammatical or disfluent). 
In our evaluation, we obtained head-to-head judgments from three workers per rewrite pair; workers agreed moderately, with a Cohen's $\kappa$=0.575 on average.
See \autoref{humaneval} for details (e.g., MTurk interface).


\subsection{Results}


Automatic metrics (\autoref{tab:auto}) show that \methodname is better at detoxification than baselines across all datasets and splits by 10.3\% on average.
Human evaluations corroborate this (\autoref{fig:human_eval}), as \methodname is on average rated as less toxic than CondBERT 2.2 times more often than vice versa across datasets and splits, and 1.9 times more often vs. ParaGeDi. 

In terms of meaning preservation as measured by BERTScore, \methodname is on par with CondBERT, with an average score within 2.5\% across datasets. However, BERTScore does not measure meaning preservation of only non-toxic content; removing toxic meaning \emph{by definition} requires trade-offs between fluency, style accuracy, and meaning preservation as discussed in most style transfer work \cite[i.a.]{dale-etal-2021-text, laugier-etal-2021-civil, malmi-etal-2020-unsupervised, ma-etal-2020-powertransformer, krishna-etal-2020-reformulating}.

Compared to DynaHate, \methodname's margin of winning is even larger on MAgr and SBF, which contain more subtle toxicity. 
For instance, in the first example from \autoref{tab:onegen}, the subtle reference to cotton picking and slavery is corrected by \methodname, which replaces ``\textit{cotton}'' with ``\textit{up}''; in contrast, both baselines fail to revise the toxic content.\footnote{Appendix \ref{rewrites} contains more example generations.} 
Since all three methods learned toxicity using the same overtly toxic data from Jigsaw, the fact that \methodname deals especially well with subtle toxicity highlights the advantages of using LMs to better model and capture toxicity patterns.

Finally, \methodname's rewrites were more fluent than other methods, according to both automatic metrics and human evaluation. \methodname's rewrites were deemed as ungrammatical the least amount of the time (9.3\%), versus 9.7\% for CondBERT and 11.7\% for ParaGeDi.

%% file: tables/human_eval.tex
\begin{figure}[t]
    \centering
    \includegraphics[width=.99\columnwidth]{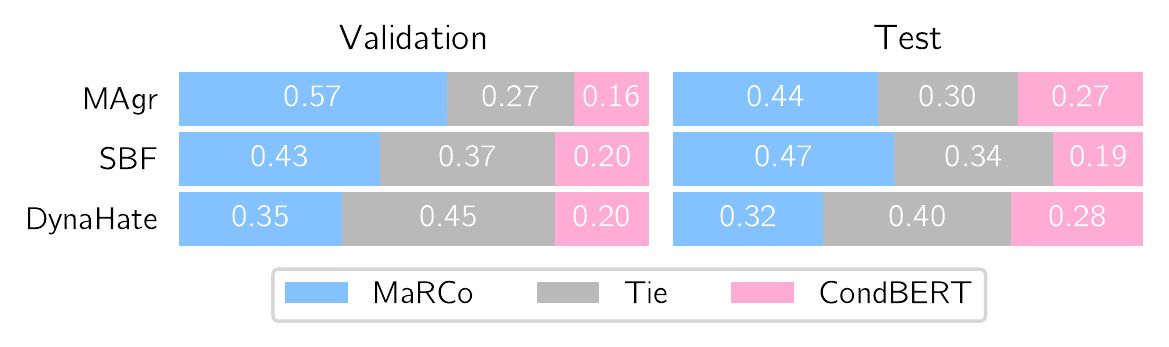}
    \includegraphics[width=.99\columnwidth]{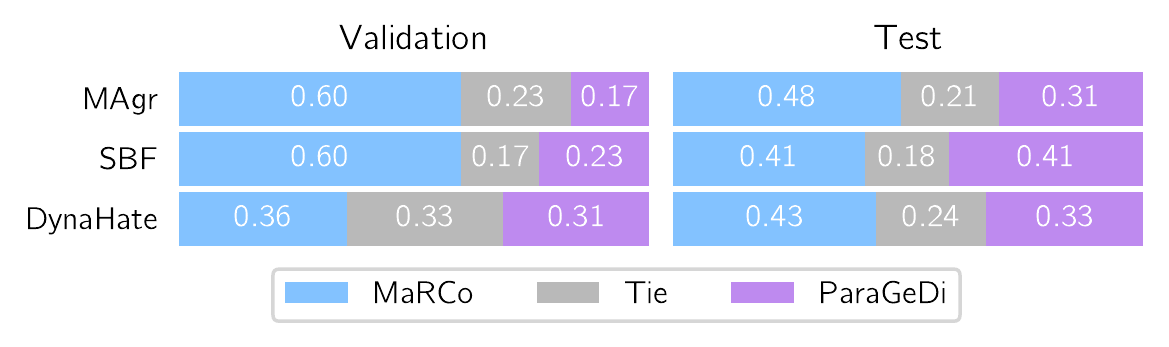}
    \caption{Head-to-head human evaluations on toxicity for \methodname vs CondBERT and \methodname vs ParaGeDi across all datasets and splits. \methodname has less-toxic generations head-to-head against both baselines, most notably on the subtle toxicity datasets (MAgr and SBF).}
    \label{fig:human_eval}
\end{figure}

%% file: tables/onegen.tex
\begin{table}[t]
\centering%
\small
\begin{tabular}{lp{5.3cm}}
\toprule 
 \rowcolor{gainsboro}\vspace{0.25em}
Original & ...because ain't nobody got time to pick all that cotton.  \\ 
 \textbf{\methodname} & ...because ain't nobody got time to pick all that up.\\
 CondBERT & ... because ain't nobody got time to pick all that cotton.  \\
 ParaGeDi & Because nobody has time to pick up all the cotton.  \\ \midrule
  \rowcolor{gainsboro}\vspace{0.25em}
 Original & Ugh, I hate living here. There's so many black people around.  \\
 \textbf{\methodname} & Ugh, I hate living here. There's so many people around. \\
CondBERT & "well, i could be living here. there's so many other nice people around.  \\
ParaGeDi & I've been living here for a long time.  \\
\midrule
   \rowcolor{gainsboro}\vspace{0.25em}
 Original & I wouldn't eat food that has been touched by a m*ng  \\
 \textbf{\methodname} & I wouldn't eat food that has been touched by a human being. \\
 CondBERT & i wouldn't be eating food that has been touched by a m*ng  \\
 ParaGeDi & I would not eat food touched by a monk.  \\
 \bottomrule
 
\end{tabular}
\caption{Different rewriting methods on three toxic examples from SBF (top), MAgr (middle), and DynaHate (bottom). In the toxic example from SBF (containing a racist slavery reference to cotton picking). \methodname detects and masks ``\textcolor{red}{cotton}'' as a toxicity indicator, which baselines fail to rewrite. 
In the last example, CondBERT fails to recognize the toxicity of the word ``m*ng'' (uncensored in the data) which is considered an ableist slur \cite{clark2011mong}.
}
\label{tab:onegen}
\end{table}

%% file: sections/06-conclusion.tex
\section{Conclusion}

We present \methodname, a novel method for text detoxification, which utilizes auto-encoder language model experts in a mask and reconstruct process.
Our method outperforms strong baselines in automatic and human evaluations, showing strong ability to detoxify even subtle biases.
\methodname's success demonstrates the effectiveness of controllable generation mixed with text rewriting methods for controllable revision, and highlights the usefulness of using LMs for capturing toxicity.


%% file: sections/07-limitations-ethics.tex
\section*{Limitations, Ethical Considerations, and Broader Impacts}
Despite the promising performance of \methodname at detoxifying text, there are several limitations, ethical considerations, and broader impacts of our approach, which we list below.


First, in this work, we seek to \emph{detoxify} sentences. However, toxicity itself is a subjective and sensitive concept with large potential downstream impacts caused by annotator and subsequent model biases \citep{sap-etal-2022-annotators}. 
We somewhat mitigate this variation by selecting human evaluators that scored highly on a toxicity qualification task (see \autoref{humaneval}), in line with a prescriptive paradigm of toxicity annotation \cite{rottger-etal-2022-two}. 
Future work could investigate the effect of demographics on preference for different rewriting algorithms, e.g., in a more descriptive paradigm.

In addition, achieving meaningful semantic preservation in detoxification is challenging. Specifically, it is difficult to disentangle the toxic and non-toxic meanings from the input, making it challenging to generate detoxified rewrites with high preservation of only the non-toxic content; this may risk minimizing marginalized groups' speech \cite{xu-etal-2021-detoxifying}. Partially, this could be due to a lack of context incorporation \cite[social, conversational, preceding sentences;][]{yerukola2023contextRewrite}; future work should consider adapting detoxification methods in context \cite{cheng2020contextual,roy2023conversation}.

\methodname also requires finetuning two pretrained LMs, which is not computationally insignificant \cite{https://doi.org/10.48550/arxiv.1906.02243, 10.1145/3381831}. Future work could explore using smaller LMs to control a larger model \cite{liu-etal-2021-dexperts}, or even more lightweight approaches.

Additionally, we acknowledge that in the evaluation, we expose Turkers to toxic content, which might harm individuals, especially those with identities that the offensive content applies to \cite{Roberts2017, 10.1145/3411764.3445092}. However, we pay a fair wage (US\$8/h) and our work is approved by our institution's ethics review board (IRB). See \autoref{humaneval} for further details.

Another major ethical implication of our work is that, following previous work, we use the Perspective API to automatically assess toxicity, a classifier which contains documented biases \cite[e.g., demographic biases and racial biases;][]{46743, sap-etal-2019-risk}. 
Future research could consider different, more holistic views of toxicity and biases \cite[e.g.,][]{sap-etal-2020-social}.

Finally, although our application in this paper is detoxification, we acknowledge that \methodname could be applied for the opposite purpose, ie., generation of toxic text from non-toxic text; this is a malicious application which we condemn. Although this issue is more prevalent for controlled generation methods \cite{DBLP:journals/corr/abs-2009-06807}, this is still a risk \methodname faces. 
In a similar vein, we do not endorse using the toxicity or microaggression datasets to develop models to generate more toxicity or microaggressions, as this may incur harm, especially to marginalized/vulnerable populations.


%% file: sections/10-appendix.tex
\section{Modeling Details}\label{modelingdetails}

\subsection{Out-of-the-Box Modeling}
We use the HuggingFace Transformers library \cite{wolf-etal-2020-transformers} version 4.10.2 for out-of-the-box, pretrained BART models and for finetuning using the \texttt{Trainer} class. It is licensed under the Apache License 2.0., and the code is available at \href{https://github.com/huggingface/transformers}{https://github.com/huggingface/transformers}.

\subsection{Finetuning the Experts}
For the expert and anti-expert models, we further finetune the base BART model with 139M parameters, found at \href{https://huggingface.co/facebook/bart-base}{https://huggingface.co/facebook/bart-base} and licensed under the Apache License 2.0, with the non-toxic and toxic corpus respectively. We use the same pretraining procedure used to further fintune BART \cite{lewis-etal-2020-bart}, and randomly corrupt sequences during training, which aligns with BART's intended use.
\paragraph{Training Corpus}
We use the Jigsaw Unintended Bias in Toxicity Classification \cite{Do2019JigsawUB} dataset for finetuning our expert and antiexpert, a corpus of forum comments on news articles. Each comment has five binary annotations on if it is toxic or not. We mark all sequences with \textbf{no} toxic annotations as \emph{non-toxic}, and all sequences with more than 50\% toxic annotations as \emph{toxic}. The intended use of this dataset is to help minimize unintended model bias, which we follow in this work.
Finally, we sample 100 instances from the validation set, and find the only individuals mentioned in Jigsaw are high-profile political figures who are already well-known. We do not perform additional anonymization of the data. 
\paragraph{Expert}
We finetune the expert with the hyperparameters listed in \autoref{tab:experthypers}, using two NVIDIA RTX6000 GPUs. We select the best checkpoint, based on the lowest evaluation loss, which is at step 100,000. The total training time is 20 hours, for 40 GPU hours of usage.

\begin{table}[H]
\footnotesize
\centering
\begin{tabular}{lr}
\toprule
          \textbf{Hyperparameter} & \textbf{Assignment}\\ \midrule
model & BART-base \\ 
number of gpus & 2\\
effective batch size & 48 \\
total steps & 100,000 \\
steps per evaluation & 1000 \\
learning rate optimizer & AdamW \\
AdamW initial learning rate & 2.5e-06 \\
AdamW epsilon & 1e-06 \\
learning rate schedule & linear with no warmup\\
weight decay & 0.0 \\
max sequence length & 180 \\
max generation length & 230 \\
padding sequences & to max seq length \\
 \bottomrule
\end{tabular}
\caption{Hyperparameters used to finetune the expert model}
\label{tab:experthypers}
\end{table}

\paragraph{Anti-Expert}
We finetune the anti-expert with the hyperparameters listed in \autoref{tab:antiexperthypers}, using a single NVIDIA RTX6000 GPU. We select the best checkpoint, based on the lowest evaluation loss, which is at step 38,000. The total training time is 2 hours, for 2 GPU hours of usage.

\begin{table}[H]
\footnotesize
\centering
\begin{tabular}{lr}
\toprule
          \textbf{Hyperparameter} & \textbf{Assignment}\\ \midrule
model & BART-base \\ 
number of gpus & 1\\
effective batch size & 32 \\
total steps & 50,000 \\
steps per evaluation & 1000 \\
learning rate optimizer & AdamW \\
AdamW initial learning rate & 1e-06 \\
AdamW epsilon & 1e-06 \\
learning rate schedule & linear with no warmup\\
weight decay & 0.0 \\
max sequence length & 180 \\
max generation length & 230 \\
padding sequences & to max seq length \\
 \bottomrule
\end{tabular}
\caption{Hyperparameters used to finetune the anti-expert model}
\label{tab:antiexperthypers}
\end{table}

\section{Experimental Details}\label{experimentaldetails}

\subsection{Datasets}
For each dataset, we manually sample and review 75 examples from the validation set, and search for any information that names or uniquely identifies individual people. We find no examples and perform no further anonymization. In addition, we follow the intended use of all three datasets by using them only to rewrite toxic sentences.

We also preprocess each of the datasets in the same way. We use the \texttt{re} package built-in to Python (we use version 3.8.11) to remove any extended white space, including tabs and line breaks, and convert them to one space. We use the \texttt{html} package, also built-in to our Python version, to convert named html character references to their corresponding string, such as ``\&gt;'' to `'>''. Afterwards, we use the \texttt{ftfy} package, version 6.1.1, found at \href{https://pypi.org/project/ftfy/}{https://pypi.org/project/ftfy/} to fix broken unicode in text. Finally, we remove any very long sequences: we calculate the 90\% percentile of text lengths to be 44, where text length is the number of space-delimited words, and we remove any sequences longer than this.

\paragraph{MAgr}
We scrape all quotes from posts using the Tumblr API, following the API License Agreement at \href{https://www.tumblr.com/docs/en/api_agreement}{https://www.tumblr.com/docs/en/api\_agreement}, which grants the right to use, distribute, display, and modify posted Tumblr content. 
\paragraph{SBF}
There is no license for this dataset.

\paragraph{DynaHate}
There is no license for this dataset.

\subsection{Generation Details}

Generations are performed using a single NVIDIA RTX6000 GPU for all datasets and methods.
\paragraph{\methodname}

\subparagraph{Masking Hyperparameters}
We set a masking threshold of $\tau=1.2$ for all experiments.

\subparagraph{Generation Hyperparameters} \label{genhypers}
We generate with greedy search for all datasets with a max generation length of 128. 

\subparagraph{MAgr}
We perform a search jointly over different hyperparameter values on the development set. We choose the hyperparameter combination that performs best on automatic metrics, shown in   \autoref{tab:magrhyper}, and use this to generate on the test set.

\begin{table}[H]
\footnotesize
\centering
\resizebox{\columnwidth}{!}{
\begin{tabular}{lrr}
\toprule
          \textbf{Hyperparameter} & \textbf{Tested} & \textbf{Assignment}\\ \midrule
 repetition penalty & [1.0, 1.2, 1.5] & 1.0 \\ 
 $\alpha_1$ & [0, 0.5, 1.0, 1.5] & 1.5 \\
 $\alpha_2$ & [3.0, 3.25, ..., 5.0] & 4.25 \\
 temperature (base model) & [0.9, 1.3, ..., 2.9] & 2.5 \\
 \bottomrule
\end{tabular}
}
\caption{Hyperparameters tested and used for \methodname on MAgr}
\label{tab:magrhyper}
\end{table}

In total, we sweep over $3 \times 4 \times 9 \times 6 = 648$ hyperparameter combinations before choosing a best set to run on our test set. Including this search, we perform approximately 150,000 rewrites. Since 100 generations take about 30 seconds, we use approximately 12.5 GPU hours.

\subparagraph{SBF}
We perform a search jointly over different hyperparameter values on the development set. We choose the hyperparameter combination that performs best on automatic metrics, shown in   \autoref{tab:sbfhyper}, and use this to generate on the test set.
\begin{table}[h]
\footnotesize
\centering
\resizebox{\columnwidth}{!}{
\begin{tabular}{lrr}
\toprule
          \textbf{Hyperparameter}& \textbf{Tested}  &  \textbf{Assignment}\\ \midrule
repetition penalty & [1.0, 1.2, 1.5] & 1.5 \\ 
 $\alpha_1$ & [0, 0.5, 1.0, 1.5] & 1.5 \\
 $\alpha_2$ & [3.0, 3.25, ..., 5.0] & 5.0 \\
temperature (base model) & [0.9, 1.3, ..., 2.9]& 2.9 \\
 \bottomrule
\end{tabular}}
\caption{Hyperparameters tested and used for \methodname on SBF}
\label{tab:sbfhyper}
\end{table}

As above, we go over $648$ hyperparameter combinations before choosing a best set to run on our test set. In total, we rewrite approximately 65,000 sequences. Since 100 generations take about 30 seconds, we use approximately 5.4 GPU hours. 

\subparagraph{DynaHate}We perform a search jointly over different hyperparameter values on the development set. We choose the hyperparameter combination that performs best on automatic metrics, shown in \autoref{tab:dynahatehyper}, and use this to generate on the test set.

\begin{table}[H]
\footnotesize
\centering
\resizebox{\columnwidth}{!}{
\begin{tabular}{lrr}
\toprule
          \textbf{Hyperparameter}& \textbf{Tested} & \textbf{Assignment}\\ \midrule
repetition penalty & [1.0, 1.2, 1.5] & 1.0 \\ 
  $\alpha_1$ & [0.5, 1.0, 1.5] & 1.5 \\
 $\alpha_2$ & [4.0, 4.25, ..., 5.0] & 4.75 \\
temperature (base model) & [0.9, 1.7, 2.5] & 2.5 \\
 \bottomrule
\end{tabular}}
\caption{Hyperparameters tested and used for \methodname on DynaHate}
\label{tab:dynahatehyper}
\end{table}
We iterate over a smaller $3 \times 3 \times 5 \times 3 = 135$ hyperparameter combinations, due to dataset size, before choosing a final set to use on our test set. In total, we rewrite approximately 240,000 texts. Since 100 generations take about 30 seconds, we use approximately 20 GPU hours.

\paragraph{Baselines}

Both of our baselines are available on \href{https://github.com/s-nlp/detox}{https://github.com/s-nlp/detox} as Jupyter Notebooks. We adapt them to Python files, runnable via the command line. There is no license available.

\subparagraph{CondBERT}

We perform a brief hyperparameter search and try two different values for the CondBERT ``number of substitute words'' hyperparameter on each validation dataset. We choose the hyperparameter that performs best on automatic metrics, given in \autoref{tab:condbert_hyper}, and use this to generate on the test sets. See \citet{dale-etal-2021-text} for a detailed description of the hyperparameter.

\begin{table}[h]
\footnotesize
\centering
\begin{tabular}{lrr}
\toprule
          \textbf{Hyperparameter} & \textbf{Tested} & \textbf{Assignment}\\ \midrule
number of substitute words & 1,10 & 1  \\ \bottomrule
\end{tabular}
\caption{Hyperparameters tested and used for CondBERT}
\label{tab:condbert_hyper}
\end{table}
Including our hyperparameter search, we run approximately 7000 rewrites across all datasets and splits. Given that 100 generations take approximately 30 seconds, our usage is 0.6 GPU hours.

CondBERT uses BERT-base, which includes 110M parameters.
\subparagraph{ParaGeDi}
We use greedy decoding for ParaGeDi and use the same hyperparameters as \methodname for each dataset, for fair comparison. Table \ref{tab:paragedihyper} lists the sole ParaGedi-specific hyperparameter we modify: we do not generate and rerank multiple sequences for fairness.

\begin{table}[H]
\footnotesize
\centering
\begin{tabular}{lr}
\toprule
          \textbf{Hyperparameter} & \textbf{Assignment}\\ \midrule
generate multiple seqs and rerank & false \\ \bottomrule
\end{tabular}
\caption{Hyperparameters used for ParaGeDi}
\label{tab:paragedihyper}
\end{table}

We perform approximately 5000 rewrites across all datasets and splits. Given that 100 generations take approximately one minute, our usage is 0.8 GPU hours.

ParaGedi uses T5-base as a paraphrasing model, with 220M parameters, in conjunction with a finetuned GPT2-medium discriminator, with 355M parameters. 
\subsection{Evaluation Metrics} \label{evalmetrics}

\paragraph{Toxicity}
To evaluate toxicity, we use the Perspective API, a publicly hosted toxicity classifier trained on the Jigsaw corpus. Given a text, the model outputs a scalar toxicity score between 0 and 1 inclusive. The model, which is located at \href{https://www.perspectiveapi.com/}{https://www.perspectiveapi.com/}, is continually updated and may change output over time. We query it in June, 2022, following the API Terms of Service and intended use at \href{https://developers.google.com/terms/}{https://developers.google.com/terms/}. 

\paragraph{Fluency}
We assess fluency by calculating the perplexity of a text with an external, pretrained language model. We use GPT2-base \cite{radford2019language}, found at \href{https://huggingface.co/gpt2}{https://huggingface.co/gpt2}, with 117M parameters, and use it under the MIT license and its intended use. 

We run this metric with a single NVIDIA RTX6000 GPU, which takes approximately 5 seconds per 100 examples. With an estimate of 450,000 texts processed, our usage for this metric is 6.3 GPU hours.

\paragraph{Meaning Preservation}
We use BERTScore \cite{DBLP:journals/corr/abs-1904-09675}, which outputs the cosine distance between model sentence embeddings, to measure the meaning similarity between the original sentence and the rewrite. We use RoBERTa-large \cite{Liu2019RoBERTaAR} as our model, which has 354M parameters. We use the code located at \href{https://huggingface.co/spaces/evaluate-metric/bertscore}{https://huggingface.co/spaces/evaluate-metric/bertscore} under the MIT License and its intended use. 

We run this evaluation with a single NVIDIA RTX6000 GPU, which takes approximately 15 seconds per 100 examples. With an estimate of 450,000 texts processed, our usage for this metric is 18.7 GPU hours.

\subsection{Total Computational Budget}
Summing up our computational usage from the above sections, including finetuning the experts, our total computational budget is 106.1 GPU hours.

\section{Example Rewrites}\label{rewrites}
Table \ref{tab:generations} shows example generations from each method across all three datasets.

\section{Human Evaluation Details}\label{humaneval}
\begin{figure*}[]
    \centering
    \includegraphics[width = 0.95\textwidth,clip,trim=0cm 0cm 0cm 3.7cm]{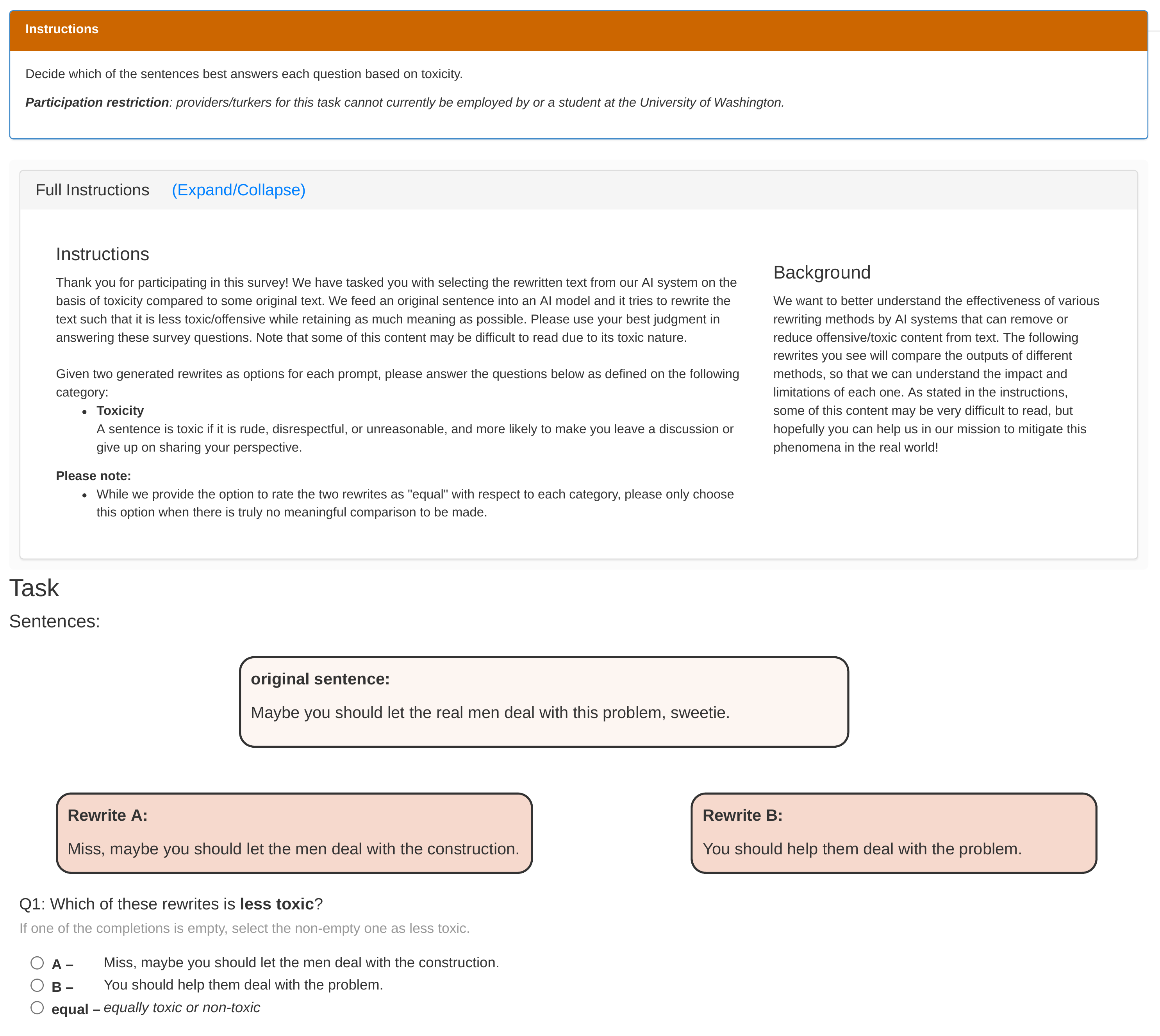}
    \caption{A screenshot of the human evaluation interface on Amazon Mechanical Turk.} 
    \label{fig:eval_ss}
\end{figure*}

We use annotators from the USA and Canada on Amazon Mechanical Turk, who voluntarily opt-in to the task. Our task was approved by our institution's ethics review board (IRB).
A screenshot of our interface for the human evaluation is shown in Figure \ref{fig:eval_ss}. Our interface describes how the annotators' data will be used. 

To gather annotations, we first recruit workers to do a qualification task, where annotators must answer six questions on which rewrite from a pair is less toxic, the same question as in our main human evaluation. The interface for this is the same as our main task shown in Figure \ref{fig:eval_ss}, but with six sentences instead of one. Annotators who answer at least five out of six questions correctly are approved and can work on the main task. We list the six examples and correct answers in \autoref{tab:quals}.

We paid a median wage of \$8/h for the qualification and the main task, which is above the minimum wage and a fair value for USA and Canada.

\section{Decoding with Product of Experts} \label{poe}
\citet{10.1162/089976602760128018} introduce the Product of Experts (PoE), an equation that states given $n$ experts: 
\begin{align}
    p(d|\theta_1, ..., \theta_n) = \frac{\prod_m p_m(d|\theta_m)}{\sum_{c} \prod_m p_m (c|\theta_m)} \label{eq:poe}
\end{align}
where $\theta_m$ denotes the parameters of model $m$, $d$ is some data vector, $p_m(d|\theta_m)$ denotes the probability of $d$ under model $m$, and $c$ iterates over all possible data vectors. 

Applying the PoE to autoregressive generation equation, $d$ represents a single token, $p_m(d|\theta_m)$ represents the next token-probability of $d$ under a specific model, and $c$ iterates over all tokens in the vocabulary $\mathcal{V}$.

Given a vector $x$, the softmax equation is:
\begin{gather}
\scalebox{0.92}{$
\begin{align*}
\text{softmax}(x_i) = \frac{e^{x_{i}}}{\sum_{j=1}^K e^{x_{j}}} \ \ \ \text{for}\ i=1,2,\dots,K
\end{align*}$}
\end{gather}
In the replacing step of \methodname, we perform the following ensembling of unnormalized log-probabilities (i.e., logits) $z_i$, $z_i^+$, and $z_i^-$ from the base and expert AE-LMs $G$, $G^+$, and $G^-$, respectively, 
conditioned on the previously generated tokens $\boldsymbol{g_{<i}}$, the original sequence $\boldsymbol{w}$, and the masked variant $\boldsymbol{w}^m$ to obtain the next-token probability:
\begin{gather}
\scalebox{0.92}{$
\begin{align*}
    P(X_i| \ \boldsymbol{g_{<i}},\boldsymbol{w}, \boldsymbol{w^m}) = \text{softmax}(z_i + \alpha_1 z_i^+ - \alpha_2 z_i^-) 
\end{align*} 
$}
\end{gather}
Consider a single index $j$ corresponding to token $\mathcal{V}_j$ in the softmax output. Given that each of $z_i$, $z_i^+$, and $z_i^-$ are \emph{log-space} probability vectors over $V$, the exponentiation of $z_i + \alpha_1 z_i^+ - \alpha_2 z_i^-$ in the softmax equation converts this sum back to a product in probability space, which matches the PoE numerator for $\mathcal{V}_j$. In addition, the denominator in the softmax equation transforms the logit sum exactly to the PoE denominator for $\mathcal{V}_j$. Thus, each probability from the output distribution from \methodname exactly matches \autoref{eq:poe} and is thus a PoE.

\input{tables/generations}
\input{tables/quals}

%% file: tables/generations.tex
\begin{table*}[]
\begin{tabular}{lll}
\toprule
\multirow{12}{*}{MAgr} & Original & ugh exams suck :'( \string#thisisgayerthanaids  \\
 & CondBERT & 'the final exams in english language:' (\string# thisisgayerthanaids  \\
& ParaGeDi & ugh exams - '(\string#thisisgayerthanaids  \\
& \methodname: & ugh exams suck :'(  \\  \cmidrule{2-3}
& Original & Your brother is such a girl!  \\
 & CondBERT & my older brother is such a lucky one!  \\
& ParaGeDi & the brother's great!  \\
 & \methodname: & Your brother is such a great guy!  \\  \cmidrule{2-3}
 & Original & Ugh, I hate living here. There's so many black people around.  \\
& CondBERT & "well, i could be living here. there's so many other nice people around.  \\
& ParaGeDi & I've been living here for a long time.  \\
 & \methodname: & Ugh, I hate living here. There's so many people around.  \\ \midrule
\multirow{12}{*}{SBF} & Original & What's the difference? All these Orientals are the same.  \\ 
 & CondBERT & what's the difference? all these orientals are the same.  \\
 & ParaGeDi & all Orientals have the same.  \\
 & \methodname: & What's the difference? All of these things are the same.  \\ \cmidrule{2-3}
 & Original & no fats no fems  \\
 & CondBERT & no demons, no demons.  \\
 & ParaGeDi & No fatty, no fats  \\
 & \methodname: & no fats no additives  \\  \cmidrule{2-3}
 & Original & ...because ain't nobody got time to pick all that cotton.  \\
 & CondBERT & ... because ain't nobody got time to pick all that cotton.  \\
 & ParaGeDi & Because nobody has time to pick up all the cotton.  \\
 & \methodname: & ...because ain't nobody got time to pick all that up.  \\ \midrule 
\multirow{12}{*}{DynaHate} & Original & I wouldn't eat food that has been touched by a mong  \\
 & CondBERT & i wouldn't be eating food that has been touched by a mong  \\
 & ParaGeDi & I would not eat food touched by a monk.  \\
 & \methodname: & I wouldn't eat food that has been touched by a human being.  \\   \cmidrule{2-3}
 & Original & I just don't like gays and that is my choice  \\
 & CondBERT & i just don't like them, and that is my choice  \\
 & ParaGeDi & I just don't like the Gays.  \\
 & \methodname: & I just don't like the idea and that is my choice  \\ \cmidrule{2-3}
 & Original & It is totally valid to say that black people are inferior  \\
 & CondBERT & it is totally valid to say that many of these people are inferior  \\
 & ParaGeDi & it's entirely true that the Black is a minority  \\
 & \methodname: & It is totally fair to say that some people are wrong. \\ \bottomrule
\end{tabular}
\caption{Examples of rewrites from each method on all datasets' validation sets}
\label{tab:generations}
\end{table*}

%% file: tables/quals.tex
\begin{table*}[]
\begin{tabular}{ll}
 \toprule
Original & i think if people don't succeed in this society, it's because they didn't work hard enough. \\
Rewrite A& if people don't succeed in this society, it's because they didn't work enough. \\
\textcolor{ForestGreen}{Rewrite B} & i think if people don't succeed in this society, it's because they didn't get lucky enough. \\
\midrule
Original & why can't we interview you? we're trying to do something for your people. \\
\textcolor{ForestGreen}{Rewrite A} & why can't we do something for you? we're trying to do something for you. \\
Rewrite B& we're trying to do something for your people. \\
\midrule
Original & i don't get it why are pregnant women working \\
\textcolor{ForestGreen}{Rewrite A}& i don't get it why are you not? \\
Rewrite B& i don't understand why pregnant women work. \\
\midrule
Original & hey honey, you're going the wrong direction, the border's that way! \\
Rewrite A & hmm... hey, honey, it's going the wrong way, the border's that way! \\
\textcolor{ForestGreen}{Rewrite B} & hey, you're going the wrong direction, the one that is! \\
\midrule
Original & When I look at you, I don’t see color \\
Rewrite A & I don't see color when I look at you. \\
\textcolor{ForestGreen}{Rewrite B}& When I look at you, I get distracted. \\
\midrule
Original & You've been doing well. It's so inspiring that you've overcome your disability! \\
\textcolor{ForestGreen}{Rewrite A}& You've been doing decently. It's so exciting! \\
Rewrite B & You've been doing well with your disability. It's so inspiring! \\
\bottomrule
\end{tabular}
\caption{The six rewrite examples used in the detoxification qualification task for workers on MTurk. The less toxic, correct rewrites are listed in \textcolor{ForestGreen}{green}.}
\label{tab:quals}
\end{table*}

%% file: 00-main.bbl
\begin{thebibliography}{46}
\expandafter\ifx\csname natexlab\endcsname\relax\def\natexlab#1{#1}\fi

\bibitem[{Breitfeller et~al.(2019)Breitfeller, Ahn, Jurgens, and
  Tsvetkov}]{breitfeller-etal-2019-finding}
Luke Breitfeller, Emily Ahn, David Jurgens, and Yulia Tsvetkov. 2019.
\newblock \href {https://doi.org/10.18653/v1/D19-1176} {Finding
  microaggressions in the wild: A case for locating elusive phenomena in social
  media posts}.
\newblock In \emph{Proceedings of the 2019 Conference on Empirical Methods in
  Natural Language Processing and the 9th International Joint Conference on
  Natural Language Processing (EMNLP-IJCNLP)}, pages 1664--1674, Hong Kong,
  China. Association for Computational Linguistics.

\bibitem[{Cheng et~al.(2020)Cheng, Gan, Zhang, Elachqar, Li, and
  Liu}]{cheng2020contextual}
Yu~Cheng, Zhe Gan, Yizhe Zhang, Oussama Elachqar, Dianqi Li, and Jingjing Liu.
  2020.
\newblock \href {https://aclanthology.org/2020.findings-emnlp.263/} {Contextual
  text style transfer}.
\newblock In \emph{Findings of the Association for Computational Linguistics:
  EMNLP 2020}, pages 2915--2924.

\bibitem[{Clark et~al.(2018)Clark, Ross, Tan, Ji, and
  Smith}]{10.1145/3172944.3172983}
Elizabeth Clark, Anne~Spencer Ross, Chenhao Tan, Yangfeng Ji, and Noah~A.
  Smith. 2018.
\newblock \href {https://doi.org/10.1145/3172944.3172983} {Creative writing
  with a machine in the loop: Case studies on slogans and stories}.
\newblock In \emph{23rd International Conference on Intelligent User
  Interfaces}, IUI '18, page 329–340, New York, NY, USA. Association for
  Computing Machinery.

\bibitem[{Clark(2011)}]{clark2011mong}
Nicola Clark. 2011.
\newblock \href
  {https://web.archive.org/web/20230405003449/https://www.theguardian.com/society/joepublic/2011/oct/19/ricky-gervais-mong-twitter}
  {Ricky gervais, please stop using the word 'mong'}.
\newblock \emph{The Gardian}.
\newblock Accessed 2023-05-25.

\bibitem[{Dale et~al.(2021)Dale, Voronov, Dementieva, Logacheva, Kozlova,
  Semenov, and Panchenko}]{dale-etal-2021-text}
David Dale, Anton Voronov, Daryna Dementieva, Varvara Logacheva, Olga Kozlova,
  Nikita Semenov, and Alexander Panchenko. 2021.
\newblock \href {https://aclanthology.org/2021.emnlp-main.629} {Text
  detoxification using large pre-trained neural models}.
\newblock In \emph{Proceedings of the 2021 Conference on Empirical Methods in
  Natural Language Processing}, pages 7979--7996, Online and Punta Cana,
  Dominican Republic. Association for Computational Linguistics.

\bibitem[{Dixon et~al.(2018)Dixon, Li, Sorensen, Thain, and Vasserman}]{46743}
Lucas Dixon, John Li, Jeffrey Sorensen, Nithum Thain, and Lucy Vasserman. 2018.
\newblock Measuring and mitigating unintended bias in text classification.

\bibitem[{Do(2019)}]{Do2019JigsawUB}
Quan~H Do. 2019.
\newblock Jigsaw unintended bias in toxicity classification.

\bibitem[{Gehman et~al.(2020)Gehman, Gururangan, Sap, Choi, and
  Smith}]{gehman-etal-2020-realtoxicityprompts}
Samuel Gehman, Suchin Gururangan, Maarten Sap, Yejin Choi, and Noah~A. Smith.
  2020.
\newblock \href {https://doi.org/10.18653/v1/2020.findings-emnlp.301}
  {{R}eal{T}oxicity{P}rompts: Evaluating neural toxic degeneration in language
  models}.
\newblock In \emph{Findings of the Association for Computational Linguistics:
  EMNLP 2020}, pages 3356--3369, Online. Association for Computational
  Linguistics.

\bibitem[{Han and Tsvetkov(2020)}]{han-tsvetkov-2020-fortifying}
Xiaochuang Han and Yulia Tsvetkov. 2020.
\newblock \href {https://doi.org/10.18653/v1/2020.emnlp-main.622} {Fortifying
  toxic speech detectors against veiled toxicity}.
\newblock In \emph{Proceedings of the 2020 Conference on Empirical Methods in
  Natural Language Processing (EMNLP)}, pages 7732--7739, Online. Association
  for Computational Linguistics.

\bibitem[{Hartvigsen et~al.(2022)Hartvigsen, Gabriel, Palangi, Sap, Ray, and
  Kamar}]{https://doi.org/10.48550/arxiv.2203.09509}
Thomas Hartvigsen, Saadia Gabriel, Hamid Palangi, Maarten Sap, Dipankar Ray,
  and Ece Kamar. 2022.
\newblock \href {https://doi.org/10.48550/ARXIV.2203.09509} {Toxigen: A
  large-scale machine-generated dataset for adversarial and implicit hate
  speech detection}.

\bibitem[{Hinton(2002)}]{10.1162/089976602760128018}
Geoffrey~E. Hinton. 2002.
\newblock \href {https://doi.org/10.1162/089976602760128018} {Training products
  of experts by minimizing contrastive divergence}.
\newblock \emph{Neural Comput.}, 14(8):1771–1800.

\bibitem[{Hohenstein et~al.(2021)Hohenstein, DiFranzo, Kizilcec, Aghajari,
  Mieczkowski, Levy, Naaman, Hancock, and Jung}]{Hohenstein2021-mn}
Jess Hohenstein, Dominic DiFranzo, Rene~F Kizilcec, Zhila Aghajari, Hannah
  Mieczkowski, Karen Levy, Mor Naaman, Jeff Hancock, and Malte Jung. 2021.
\newblock \href {http://arxiv.org/abs/2102.05756} {Artificial intelligence in
  communication impacts language and social relationships}.

\bibitem[{Hu et~al.(2017)Hu, Yang, Liang, Salakhutdinov, and
  Xing}]{10.5555/3305381.3305545}
Zhiting Hu, Zichao Yang, Xiaodan Liang, Ruslan Salakhutdinov, and Eric~P. Xing.
  2017.
\newblock Toward controlled generation of text.
\newblock In \emph{Proceedings of the 34th International Conference on Machine
  Learning - Volume 70}, ICML'17, page 1587–1596. JMLR.org.

\bibitem[{Jhamtani et~al.(2017)Jhamtani, Gangal, Hovy, and
  Nyberg}]{jhamtani-etal-2017-shakespearizing}
Harsh Jhamtani, Varun Gangal, Eduard Hovy, and Eric Nyberg. 2017.
\newblock \href {https://doi.org/10.18653/v1/W17-4902} {Shakespearizing modern
  language using copy-enriched sequence to sequence models}.
\newblock In \emph{Proceedings of the Workshop on Stylistic Variation}, pages
  10--19, Copenhagen, Denmark. Association for Computational Linguistics.

\bibitem[{Kiritchenko and Mohammad(2017)}]{kiritchenko-mohammad-2017-best}
Svetlana Kiritchenko and Saif Mohammad. 2017.
\newblock \href {https://doi.org/10.18653/v1/P17-2074} {Best-worst scaling more
  reliable than rating scales: A case study on sentiment intensity annotation}.
\newblock In \emph{Proceedings of the 55th Annual Meeting of the Association
  for Computational Linguistics (Volume 2: Short Papers)}, pages 465--470,
  Vancouver, Canada. Association for Computational Linguistics.

\bibitem[{Krishna et~al.(2020)Krishna, Wieting, and
  Iyyer}]{krishna-etal-2020-reformulating}
Kalpesh Krishna, John Wieting, and Mohit Iyyer. 2020.
\newblock \href {https://doi.org/10.18653/v1/2020.emnlp-main.55} {Reformulating
  unsupervised style transfer as paraphrase generation}.
\newblock In \emph{Proceedings of the 2020 Conference on Empirical Methods in
  Natural Language Processing (EMNLP)}, pages 737--762, Online. Association for
  Computational Linguistics.

\bibitem[{Laugier et~al.(2021)Laugier, Pavlopoulos, Sorensen, and
  Dixon}]{laugier-etal-2021-civil}
L{\'e}o Laugier, John Pavlopoulos, Jeffrey Sorensen, and Lucas Dixon. 2021.
\newblock \href {https://doi.org/10.18653/v1/2021.eacl-main.124} {Civil
  rephrases of toxic texts with self-supervised transformers}.
\newblock In \emph{Proceedings of the 16th Conference of the European Chapter
  of the Association for Computational Linguistics: Main Volume}, pages
  1442--1461, Online. Association for Computational Linguistics.

\bibitem[{Lewis et~al.(2020)Lewis, Liu, Goyal, Ghazvininejad, Mohamed, Levy,
  Stoyanov, and Zettlemoyer}]{lewis-etal-2020-bart}
Mike Lewis, Yinhan Liu, Naman Goyal, Marjan Ghazvininejad, Abdelrahman Mohamed,
  Omer Levy, Veselin Stoyanov, and Luke Zettlemoyer. 2020.
\newblock \href {https://doi.org/10.18653/v1/2020.acl-main.703} {{BART}:
  Denoising sequence-to-sequence pre-training for natural language generation,
  translation, and comprehension}.
\newblock In \emph{Proceedings of the 58th Annual Meeting of the Association
  for Computational Linguistics}, pages 7871--7880, Online. Association for
  Computational Linguistics.

\bibitem[{Li et~al.(2018)Li, Jia, He, and Liang}]{li-etal-2018-delete}
Juncen Li, Robin Jia, He~He, and Percy Liang. 2018.
\newblock \href {https://doi.org/10.18653/v1/N18-1169} {Delete, retrieve,
  generate: a simple approach to sentiment and style transfer}.
\newblock In \emph{Proceedings of the 2018 Conference of the North {A}merican
  Chapter of the Association for Computational Linguistics: Human Language
  Technologies, Volume 1 (Long Papers)}, pages 1865--1874, New Orleans,
  Louisiana. Association for Computational Linguistics.

\bibitem[{Liu et~al.(2021)Liu, Sap, Lu, Swayamdipta, Bhagavatula, Smith, and
  Choi}]{liu-etal-2021-dexperts}
Alisa Liu, Maarten Sap, Ximing Lu, Swabha Swayamdipta, Chandra Bhagavatula,
  Noah~A. Smith, and Yejin Choi. 2021.
\newblock \href {https://doi.org/10.18653/v1/2021.acl-long.522} {{DE}xperts:
  Decoding-time controlled text generation with experts and anti-experts}.
\newblock In \emph{Proceedings of the 59th Annual Meeting of the Association
  for Computational Linguistics and the 11th International Joint Conference on
  Natural Language Processing (Volume 1: Long Papers)}, pages 6691--6706,
  Online. Association for Computational Linguistics.

\bibitem[{Liu et~al.(2019)Liu, Ott, Goyal, Du, Joshi, Chen, Levy, Lewis,
  Zettlemoyer, and Stoyanov}]{Liu2019RoBERTaAR}
Yinhan Liu, Myle Ott, Naman Goyal, Jingfei Du, Mandar Joshi, Danqi Chen, Omer
  Levy, Mike Lewis, Luke Zettlemoyer, and Veselin Stoyanov. 2019.
\newblock Roberta: A robustly optimized bert pretraining approach.
\newblock \emph{ArXiv}, abs/1907.11692.

\bibitem[{Ma et~al.(2020)Ma, Sap, Rashkin, and
  Choi}]{ma-etal-2020-powertransformer}
Xinyao Ma, Maarten Sap, Hannah Rashkin, and Yejin Choi. 2020.
\newblock \href {https://doi.org/10.18653/v1/2020.emnlp-main.602}
  {{P}ower{T}ransformer: Unsupervised controllable revision for biased language
  correction}.
\newblock In \emph{Proceedings of the 2020 Conference on Empirical Methods in
  Natural Language Processing (EMNLP)}, pages 7426--7441, Online. Association
  for Computational Linguistics.

\bibitem[{Malmi et~al.(2020)Malmi, Severyn, and
  Rothe}]{malmi-etal-2020-unsupervised}
Eric Malmi, Aliaksei Severyn, and Sascha Rothe. 2020.
\newblock \href {https://doi.org/10.18653/v1/2020.emnlp-main.699} {Unsupervised
  text style transfer with padded masked language models}.
\newblock In \emph{Proceedings of the 2020 Conference on Empirical Methods in
  Natural Language Processing (EMNLP)}, pages 8671--8680, Online. Association
  for Computational Linguistics.

\bibitem[{McGuffie and Newhouse(2020)}]{DBLP:journals/corr/abs-2009-06807}
Kris McGuffie and Alex Newhouse. 2020.
\newblock \href {http://arxiv.org/abs/2009.06807} {The radicalization risks of
  {GPT-3} and advanced neural language models}.
\newblock \emph{CoRR}, abs/2009.06807.

\bibitem[{Nadal et~al.(2014)Nadal, Griffin, Wong, Hamit, and
  Rasmus}]{https://doi.org/10.1002/j.1556-6676.2014.00130.x}
Kevin~L. Nadal, Katie~E. Griffin, Yinglee Wong, Sahran Hamit, and Morgan
  Rasmus. 2014.
\newblock \href
  {https://doi.org/https://doi.org/10.1002/j.1556-6676.2014.00130.x} {The
  impact of racial microaggressions on mental health: Counseling implications
  for clients of color}.
\newblock \emph{Journal of Counseling \& Development}, 92(1):57--66.

\bibitem[{Nogueira~dos Santos et~al.(2018)Nogueira~dos Santos, Melnyk, and
  Padhi}]{nogueira-dos-santos-etal-2018-fighting}
Cicero Nogueira~dos Santos, Igor Melnyk, and Inkit Padhi. 2018.
\newblock \href {https://doi.org/10.18653/v1/P18-2031} {Fighting offensive
  language on social media with unsupervised text style transfer}.
\newblock In \emph{Proceedings of the 56th Annual Meeting of the Association
  for Computational Linguistics (Volume 2: Short Papers)}, pages 189--194,
  Melbourne, Australia. Association for Computational Linguistics.

\bibitem[{{OHCHR}(2021)}]{Ohchr2021-ba}
{OHCHR}. 2021.
\newblock \href
  {https://www.ohchr.org/en/stories/2021/03/report-online-hate-increasing-against-minorities-says-expert}
  {Report: Online hate increasing against minorities, says expert}.
\newblock Technical report.

\bibitem[{Prabhumoye et~al.(2020)Prabhumoye, Black, and
  Salakhutdinov}]{prabhumoye-etal-2020-exploring}
Shrimai Prabhumoye, Alan~W Black, and Ruslan Salakhutdinov. 2020.
\newblock \href {https://doi.org/10.18653/v1/2020.coling-main.1} {Exploring
  controllable text generation techniques}.
\newblock In \emph{Proceedings of the 28th International Conference on
  Computational Linguistics}, pages 1--14, Barcelona, Spain (Online).
  International Committee on Computational Linguistics.

\bibitem[{Radford et~al.(2019)Radford, Wu, Child, Luan, Amodei, Sutskever
  et~al.}]{radford2019language}
Alec Radford, Jeffrey Wu, Rewon Child, David Luan, Dario Amodei, Ilya
  Sutskever, et~al. 2019.
\newblock Language models are unsupervised multitask learners.
\newblock \emph{OpenAI blog}, 1(8):9.

\bibitem[{Roberts(2017)}]{Roberts2017}
Sarah~T Roberts. 2017.
\newblock \href
  {https://www.theatlantic.com/technology/archive/2017/03/commercial-content-moderation/518796/}
  {Social media's silent filter}.
\newblock \emph{The Atlantic}.

\bibitem[{Rottger et~al.(2022)Rottger, Vidgen, Hovy, and
  Pierrehumbert}]{rottger-etal-2022-two}
Paul Rottger, Bertie Vidgen, Dirk Hovy, and Janet Pierrehumbert. 2022.
\newblock \href {https://doi.org/10.18653/v1/2022.naacl-main.13} {Two
  contrasting data annotation paradigms for subjective {NLP} tasks}.
\newblock In \emph{Proceedings of the 2022 Conference of the North American
  Chapter of the Association for Computational Linguistics: Human Language
  Technologies}, pages 175--190, Seattle, United States. Association for
  Computational Linguistics.

\bibitem[{Roy et~al.(2023)Roy, Shu, Pappas, Mansimov, Zhang, Mansour, and
  Roth}]{roy2023conversation}
Shamik Roy, Raphael Shu, Nikolaos Pappas, Elman Mansimov, Yi~Zhang, Saab
  Mansour, and Dan Roth. 2023.
\newblock Conversation style transfer using few-shot learning.
\newblock \emph{arXiv preprint arXiv:2302.08362}.

\bibitem[{Sap et~al.(2019)Sap, Card, Gabriel, Choi, and
  Smith}]{sap-etal-2019-risk}
Maarten Sap, Dallas Card, Saadia Gabriel, Yejin Choi, and Noah~A. Smith. 2019.
\newblock \href {https://doi.org/10.18653/v1/P19-1163} {The risk of racial bias
  in hate speech detection}.
\newblock In \emph{Proceedings of the 57th Annual Meeting of the Association
  for Computational Linguistics}, pages 1668--1678, Florence, Italy.
  Association for Computational Linguistics.

\bibitem[{Sap et~al.(2020)Sap, Gabriel, Qin, Jurafsky, Smith, and
  Choi}]{sap-etal-2020-social}
Maarten Sap, Saadia Gabriel, Lianhui Qin, Dan Jurafsky, Noah~A. Smith, and
  Yejin Choi. 2020.
\newblock \href {https://doi.org/10.18653/v1/2020.acl-main.486} {Social bias
  frames: Reasoning about social and power implications of language}.
\newblock In \emph{Proceedings of the 58th Annual Meeting of the Association
  for Computational Linguistics}, pages 5477--5490, Online. Association for
  Computational Linguistics.

\bibitem[{Sap et~al.(2022)Sap, Swayamdipta, Vianna, Zhou, Choi, and
  Smith}]{sap-etal-2022-annotators}
Maarten Sap, Swabha Swayamdipta, Laura Vianna, Xuhui Zhou, Yejin Choi, and Noah
  Smith. 2022.
\newblock \href {https://aclanthology.org/2022.naacl-main.431} {Annotators with
  attitudes: How annotator beliefs and identities bias toxic language
  detection}.
\newblock In \emph{Proceedings of the 2022 Conference of the North American
  Chapter of the Association for Computational Linguistics: Human Language
  Technologies}, pages 5884--5906, Seattle, United States. Association for
  Computational Linguistics.

\bibitem[{Schwartz et~al.(2020)Schwartz, Dodge, Smith, and
  Etzioni}]{10.1145/3381831}
Roy Schwartz, Jesse Dodge, Noah~A. Smith, and Oren Etzioni. 2020.
\newblock \href {https://doi.org/10.1145/3381831} {Green ai}.
\newblock \emph{Commun. ACM}, 63(12):54–63.

\bibitem[{Shen et~al.(2017)Shen, Lei, Barzilay, and
  Jaakkola}]{10.5555/3295222.3295427}
Tianxiao Shen, Tao Lei, Regina Barzilay, and Tommi Jaakkola. 2017.
\newblock Style transfer from non-parallel text by cross-alignment.
\newblock In \emph{Proceedings of the 31st International Conference on Neural
  Information Processing Systems}, NIPS'17, page 6833–6844, Red Hook, NY,
  USA. Curran Associates Inc.

\bibitem[{Steiger et~al.(2021)Steiger, Bharucha, Venkatagiri, Riedl, and
  Lease}]{10.1145/3411764.3445092}
Miriah Steiger, Timir~J Bharucha, Sukrit Venkatagiri, Martin~J. Riedl, and
  Matthew Lease. 2021.
\newblock \href {https://doi.org/10.1145/3411764.3445092} {The psychological
  well-being of content moderators: The emotional labor of commercial
  moderation and avenues for improving support}.
\newblock In \emph{Proceedings of the 2021 CHI Conference on Human Factors in
  Computing Systems}, CHI '21, New York, NY, USA. Association for Computing
  Machinery.

\bibitem[{Strubell et~al.(2019)Strubell, Ganesh, and
  McCallum}]{https://doi.org/10.48550/arxiv.1906.02243}
Emma Strubell, Ananya Ganesh, and Andrew McCallum. 2019.
\newblock \href {https://doi.org/10.48550/ARXIV.1906.02243} {Energy and policy
  considerations for deep learning in nlp}.

\bibitem[{Thomas et~al.(2021)Thomas, Akhawe, Bailey, Boneh, Bursztein,
  Consolvo, Dell, Durumeric, Kelley, Kumar, McCoy, Meiklejohn, Ristenpart, and
  Stringhini}]{9519435}
Kurt Thomas, Devdatta Akhawe, Michael Bailey, Dan Boneh, Elie Bursztein, Sunny
  Consolvo, Nicola Dell, Zakir Durumeric, Patrick~Gage Kelley, Deepak Kumar,
  Damon McCoy, Sarah Meiklejohn, Thomas Ristenpart, and Gianluca Stringhini.
  2021.
\newblock \href {https://doi.org/10.1109/SP40001.2021.00028} {Sok: Hate,
  harassment, and the changing landscape of online abuse}.
\newblock In \emph{2021 IEEE Symposium on Security and Privacy (SP)}, pages
  247--267.

\bibitem[{Vidgen et~al.(2021)Vidgen, Thrush, Waseem, and
  Kiela}]{vidgen-etal-2021-learning}
Bertie Vidgen, Tristan Thrush, Zeerak Waseem, and Douwe Kiela. 2021.
\newblock \href {https://doi.org/10.18653/v1/2021.acl-long.132} {Learning from
  the worst: Dynamically generated datasets to improve online hate detection}.
\newblock In \emph{Proceedings of the 59th Annual Meeting of the Association
  for Computational Linguistics and the 11th International Joint Conference on
  Natural Language Processing (Volume 1: Long Papers)}, pages 1667--1682,
  Online. Association for Computational Linguistics.

\bibitem[{Wolf et~al.(2020)Wolf, Debut, Sanh, Chaumond, Delangue, Moi, Cistac,
  Rault, Louf, Funtowicz, Davison, Shleifer, von Platen, Ma, Jernite, Plu, Xu,
  Le~Scao, Gugger, Drame, Lhoest, and Rush}]{wolf-etal-2020-transformers}
Thomas Wolf, Lysandre Debut, Victor Sanh, Julien Chaumond, Clement Delangue,
  Anthony Moi, Pierric Cistac, Tim Rault, Remi Louf, Morgan Funtowicz, Joe
  Davison, Sam Shleifer, Patrick von Platen, Clara Ma, Yacine Jernite, Julien
  Plu, Canwen Xu, Teven Le~Scao, Sylvain Gugger, Mariama Drame, Quentin Lhoest,
  and Alexander Rush. 2020.
\newblock \href {https://doi.org/10.18653/v1/2020.emnlp-demos.6} {Transformers:
  State-of-the-art natural language processing}.
\newblock In \emph{Proceedings of the 2020 Conference on Empirical Methods in
  Natural Language Processing: System Demonstrations}, pages 38--45, Online.
  Association for Computational Linguistics.

\bibitem[{Wu et~al.(2019)Wu, Zhang, Zang, Han, and Hu}]{ijcai2019-732}
Xing Wu, Tao Zhang, Liangjun Zang, Jizhong Han, and Songlin Hu. 2019.
\newblock \href {https://doi.org/10.24963/ijcai.2019/732} {Mask and infill:
  Applying masked language model for sentiment transfer}.
\newblock In \emph{Proceedings of the Twenty-Eighth International Joint
  Conference on Artificial Intelligence, {IJCAI-19}}, pages 5271--5277.
  International Joint Conferences on Artificial Intelligence Organization.

\bibitem[{Xu et~al.(2021)Xu, Pathak, Wallace, Gururangan, Sap, and
  Klein}]{xu-etal-2021-detoxifying}
Albert Xu, Eshaan Pathak, Eric Wallace, Suchin Gururangan, Maarten Sap, and Dan
  Klein. 2021.
\newblock \href {https://doi.org/10.18653/v1/2021.naacl-main.190} {Detoxifying
  language models risks marginalizing minority voices}.
\newblock In \emph{Proceedings of the 2021 Conference of the North American
  Chapter of the Association for Computational Linguistics: Human Language
  Technologies}, pages 2390--2397, Online. Association for Computational
  Linguistics.

\bibitem[{Yerukola et~al.(2023)Yerukola, Zhou, and
  Sap}]{yerukola2023contextRewrite}
Akhila Yerukola, Xuhui Zhou, and Maarten Sap. 2023.
\newblock ``don't take this out of context!'' on the need for contextual models
  and evaluations for stylistic rewriting.
\newblock \emph{arXiv}.

\bibitem[{Zhang et~al.(2019)Zhang, Kishore, Wu, Weinberger, and
  Artzi}]{DBLP:journals/corr/abs-1904-09675}
Tianyi Zhang, Varsha Kishore, Felix Wu, Kilian~Q. Weinberger, and Yoav Artzi.
  2019.
\newblock \href {http://arxiv.org/abs/1904.09675} {Bertscore: Evaluating text
  generation with {BERT}}.
\newblock \emph{CoRR}, abs/1904.09675.

\end{thebibliography}
